\documentclass[runningheads]{llncs}
\usepackage{graphicx}
\usepackage{amsmath,amssymb} 
\usepackage{color}
\usepackage{hyperref}

\usepackage{multirow}
\usepackage{subcaption}
\usepackage{tablefootnote}
\usepackage[dvipsnames]{xcolor}
\usepackage[symbol]{footmisc}

\def\eg{\emph{e.g.\ }}
\def\ie{\emph{i.e.\ }} 
\def\etal{\emph{et al.\ }}

\begin{document}
\title{Action Search: Spotting Actions in Videos and Its Application to Temporal Action Localization} 

\titlerunning{Action Search: Spotting Actions in Videos}
%
\author{Humam Alwassel \and
Fabian Caba Heilbron \and
Bernard Ghanem}
%
\authorrunning{H. Alwassel \etal}
%

\institute{King Abdullah University of Science and Technology (KAUST), Saudi Arabia\\
\url{http://www.humamalwassel.com/publication/action-search/}
\email{\{humam.alwassel,fabian.caba,bernard.ghanem\}@kaust.edu.sa}
}
\maketitle              
\begin{abstract}
    State-of-the-art temporal action detectors inefficiently search the entire video for specific actions. Despite the encouraging progress these methods achieve, it is crucial to design automated approaches that only explore parts of the video which are the most relevant to the actions being searched for. To address this need, we propose the new problem of \emph{action spotting} in video, which we define as finding a specific action in a video while observing a small portion of that video. Inspired by the observation that humans are extremely efficient and accurate in spotting and finding action instances in video, we propose \emph{Action Search}, a novel Recurrent Neural Network approach that mimics the way humans spot actions. Moreover, to address the absence of data recording the behavior of human annotators, we put forward the \emph{Human Searches} dataset, which compiles the search sequences employed by human annotators spotting actions in the AVA and THUMOS14 datasets. We consider temporal action localization as an application of the \emph{action spotting} problem. Experiments on the THUMOS14 dataset reveal that our model is not only able to explore the video efficiently (observing on average $\mathbf{17.3}\%$ of the video) but it also accurately finds human activities with $\mathbf{30.8}\%$ mAP.{\let\thefootnote\relax\footnote{{The first two authors contributed equally to this work. Authors ordering was determined by three coin flips.}}}
    \keywords{Video understanding $\cdot$ Action localization $\cdot$ Action spotting}
\end{abstract}

\section{Introduction}
Similar to many video-related applications, such as video object detection and video surveillance, temporal action localization requires an efficient search for different visual targets in videos. With the recent exponential growth in the number videos online (\eg over 400 video hours are uploaded to YouTube every minute), it is crucial today to develop methods that can simultaneously search this large volume of videos efficiently and spot actions accurately. Thus, we propose the new problem of \emph{action spotting} in video, which we define as finding a specific action in a video sequence while observing a small portion of that video. Since the computational cost is directly impacted by the number of observations made in a video, this spotting problem brings search efficiency to the forefront. Obviously, the overall computational cost of such an action search can be reduced by making the per-observation computation faster, as done in many previous work; however, as the video becomes long and the action instances sparse in the video, the number of observations needed dominates this cost. 

\begin{figure}[!t]
    \centering
    \begin{subfigure}{0.53\linewidth}
        \includegraphics[width=1\textwidth]{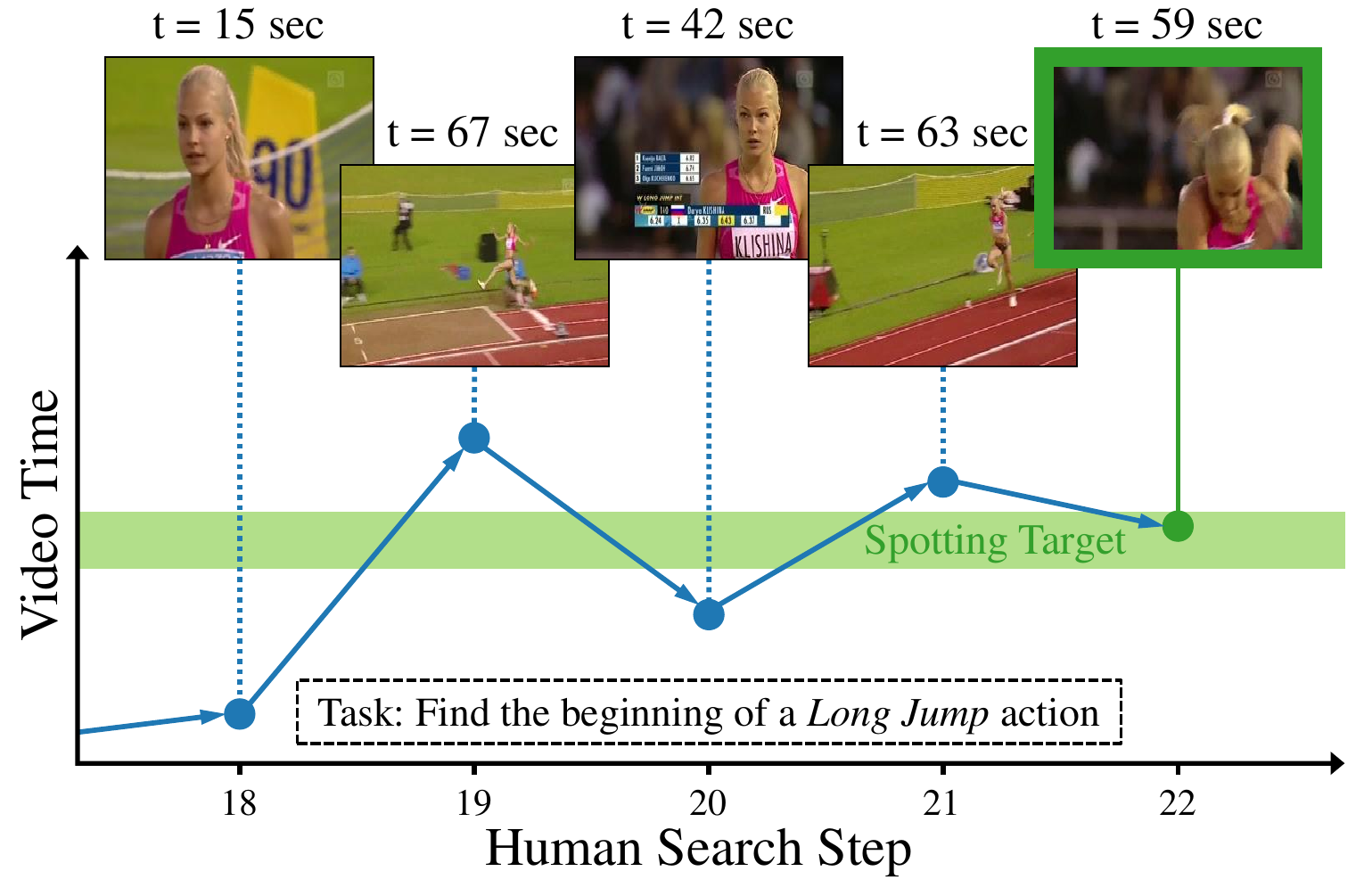}
    \end{subfigure}
    \begin{subfigure}{0.39\linewidth}
        \includegraphics[width=1\textwidth]{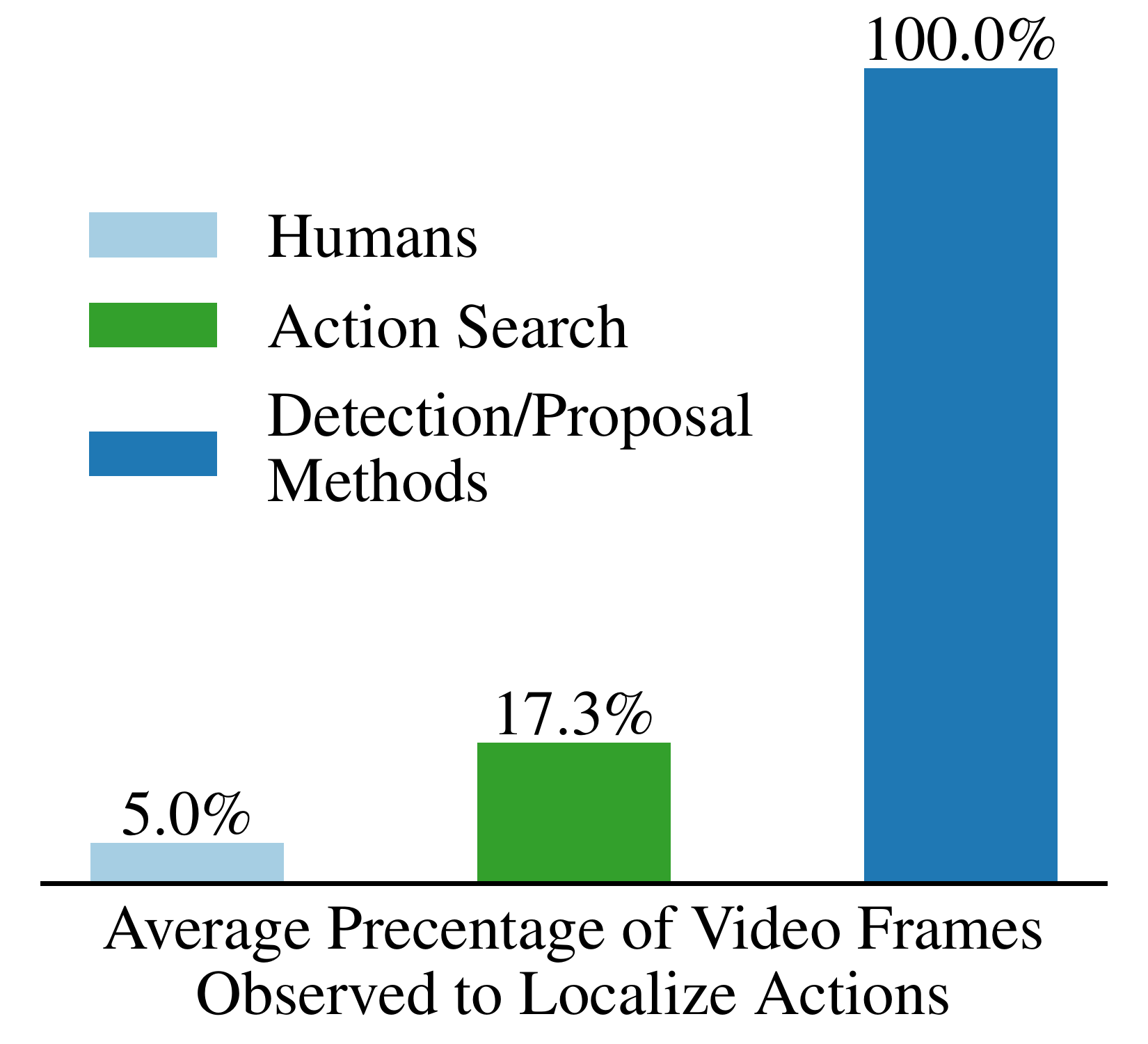}
    \end{subfigure}
    \caption{
        \textbf{Left:} A partial search sequence a human performed to find the start of a \emph{Long Jump} action (the shaded green area). Notably, humans are efficient in spotting actions without observing a large portion of the video. \textbf{Right:} An efficiency comparison between humans, \emph{Action Search}, and other detection/proposal methods on THUMOS14 \cite{thumos14}. Our model is $5.8$x more efficient than other methods.
    }
    \label{fig:pull}
\end{figure}

To get intuitions on how to automatically perform an efficient action search in video, we take notice of how humans approach the problem. In Figure \ref{fig:pull} (left), we show part of a search sequence a human observer carries out when asked to find the beginning of a \emph{Long Jump} action in a long video. This sequence reveals that the person can quickly find the spotting target (in $22$ search steps) without observing the entire video, which indicates the possible role temporal context plays in searching for actions. In this case, only a very small portion of the video is observed before the search successfully terminates. In fact, we observe a similar search pattern for different action targets and across different annotators. 

Early approaches for temporal action localization rely on trimmed data to learn sophisticated models \cite{oneata_cvpr2014,res3d,ssn}. These methods achieve great success at detecting human actions. Despite the encouraging progress in action localization and the attention it is recently attracting, the goal of accurate detection remains elusive to automated systems. A main drawback of current detection methods is that they do not exploit (and totally discard) the search process human annotators follow to produce the final temporal annotations, which are inherently the only information from the annotation process used to train these detection models. As such, these methods need to scan the whole video in an exhaustive manner (either by observing all video frames or a uniform temporal subsampling of them) to detect human actions. This is inefficient; thus, it is crucial to investigate methods that observe the smallest percentage of the video and maintain a state-of-the-art mAP. This naturally leads to faster, more efficient methods, which implies that \emph{action spotting} is essential for temporal action localization.

In this paper, we focus on \emph{action spotting} as a precursor to temporal action localization, since one can reformulate action localization as an \emph{action spotting} problem
followed by a regression task to define the action length. Based on observations drawn from the user study in Section \ref{section:action_searches}, we believe that \emph{action spotting} is an effective precursor that departs from the focus of the traditional action proposal precursor. 
Action proposal generation aims to localize the temporal bounds of actions as tightly as possible, while the goal of \emph{action spotting} is to search for action instances as efficiently as possible.
Moreover, action proposal generation is class-agnostic, while \emph{action spotting} is class-specific. Figure \ref{fig:pull} (right) compares the efficiency of action localization using traditional approaches (\eg proposals) against using our \emph{action spotting}-based method.

Recent temporal action localization datasets \cite{activitynet,thumos14,charades}
use human annotators to label the action boundaries in a video. Although these datasets are opening new and exciting challenges in the field, they lack an important component: the sequence of steps the human annotator follows to produce the final annotation. These search sequences can be collected for \emph{free} when creating new datasets or extending current ones, and they are a valuable resource for \emph{action spotting}. 

Inspired by the observation that humans are extremely efficient and accurate in finding individual action instances in a video, we aim to solve the \emph{action spotting} problem in this paper by  imitating how humans search in videos. 

\noindent \textbf{Contributions.} 
\textbf{(i)} To address the lack of data on the behavior of human annotators, we put forward the \emph{Human Searches} dataset, a new dataset composed of the search sequences of human annotators for the AVA \cite{ava} and THUMOS14 \cite{thumos14} datasets (Section \ref{section:action_searches}).
\textbf{(ii)} We propose \emph{Action Search}, a novel Recurrent Neural Network approach that mimics the way humans spot actions in untrimmed videos (Section \ref{section:model}). 
\textbf{(iii)} We validate \emph{Action Search} in the \emph{action spotting} problem by demonstrating it requires on average $\mathbf{16.6}\%$ and $\mathbf{22.3}\%$ fewer observations to successfully spot an action than two baseline models (Section \ref{section:exp_spotting}). Moreover, when our model is used in the domain of temporal action localization, it achieves state-of-the-art detection results on the THUMOS14 \cite{thumos14} dataset with $\mathbf{30.8\%}$ mAP while only observing on average $\mathbf{17.3\%}$ of the video (Section \ref{section:exp_localization}).

\section{Related Work}

\noindent\textbf{Datasets.} Recognizing and localizing human activities in video often require an extensive
collection of annotated data. In recent years, several datasets for temporal 
action localization have become available. For instance, Jiang \etal \cite{thumos14} introduce THUMOS14, a large-scale dataset of untrimmed video sequences with 
$20$ different sports categories. Concurrently, ActivityNet \cite{activitynet} 
establishes a large benchmark of long YouTube videos with $200$ annotated daily activities. Later, Sigurdsson \etal \cite{charades} release \emph{Charades}, a day-to-day indoor actions database captured in a crowdsourced manner.
More recently, Google introduced AVA \cite{ava}, short for \emph{atomic visual actions}, a densely annotated dataset localizing human actions in space and time. All four datasets use human annotators to localize intended activities in a video. 
Although these datasets are opening new challenges in the field, they all miss an important component: the search sequence the human annotator follows to produce the final annotation, which is a form of supervised annotation that can be collected for \emph{free} during the annotation process. In Section \ref{section:action_searches}, we introduce the \emph{Human Searches} dataset,  which allows us to disrupt the current paradigm on how action datasets are structured/collected and how action spotting/localization models can potentially be trained.

\noindent\textbf{Temporal Action Localization.}
A large number of works have successfully tackled the task of action recognition \cite{caba_accv14,quo_vadis,niebles_eccv2010,simonyan_nips2014,densetraj} and spatio-temporal localization \cite{actionness,actiontubes,spoton,peng_eccv2016,saha_bmvc2016,soomro_cvpr2016}. 
Here, we briefly review some of the influential works in the realm of temporal action localization. Early methods have relied on the sliding-window-plus-classifier combination to produce the actions temporal boundaries \cite{duchenne_iccv2009,gaidon_cvpr2011,jain_cvpr2015,oneata_cvpr2014}.
Recently, a series of works have explored the idea of action proposals to reduce the computational complexity incurred by sliding window-based approaches.
Notably, Shou \etal \cite{scnn} introduce a multi-stage system that finds and classifies interest regions to produce temporal action locations. Meanwhile, Caba Heilbron \etal \cite{sparseprop} propose a sparse learning framework to rank a segment based on its similarity to training samples. In the same spirit, contemporary works \cite{sst,daps,turntap} produce action proposals by exploiting the effectiveness of deep neural networks. More recently, end-to-end methods have proven to boost the performance of two-stage approaches, demonstrating the importance of jointly optimizing the feature extraction and detection process \cite{sstad,res3d,rc3d,ssn}. Other researchers have used language models \cite{richard_cvpr2016}, action progression analysis \cite{shugao_cvpr2016,cdc}, and high-level semantics \cite{scc} to produce high-fidelity action detections.

The large body of work on temporal action localization has mostly focused on improvements in detection performance and/or speed, while very few works have targeted the development of efficient search mechanisms. The dominant search strategy in prior work has focused on exhaustively observing the entire video (either by observing all frames or a uniform temporal subsampling of them) at least once. The pioneering approach of Yeung \etal \cite{frameglimpses} comes as a departure from this paradigm, whereby they introduce the \emph{Frame Glimpses} method to predict the temporal bounds of an action by observing very few frames. Their reinforcement learning-based method tries to learn a policy to intelligently jump through the video for the immediate purpose of detection. While it shares a similar goal, our model avoids the subtleties of learning such policies by exploiting ground truth data depicting how humans sequentially annotate actions. Using such information for action localization, our method outperforms \emph{Frame Glimpses} and achieves state-of-the-art detection results on THUMOS14 \cite{thumos14} (see Section \ref{section:exp_localization}).

\noindent\textbf{Sequence Prediction.}
Recurrent Neural Networks (RNNs) and specifically Long Short Term Memory (LSTM) networks have successfully tackled several sequence prediction problems \cite{alahi_cvpr2016,bengio_nips2015,graves_arxiv2013,graves_icml2014}. 
For instance, Alahi \etal \cite{alahi_cvpr2016} introduce an LSTM based model to predict human motion trajectories in crowded scenes. Graves \etal \cite{graves_arxiv2013} propose an RNN that learns to predict the 
next stroke in online handwriting prediction. Motivated by the success of  these approaches, Section \ref{section:model} introduces our novel Learning-to-Search strategy, which formulates the problem of \emph{action spotting} as a sequence prediction problem.

\section{Action Spotting: What do Humans do?} \label{section:action_searches}
In this section and motivated by how well humans spot and localize actions in videos, we investigate some factors that intuitively seem to play a role in this process. Specifically, we address the following questions: are humans distracted when they are asked to find multiple action classes? and is spotting an action easier than finding its temporal boundaries? Finally, we describe our key idea.

\noindent\textbf{Single vs. Multiple Class Search.} To investigate whether humans are distracted when asked to find multiple action classes, we conduct an online user study on Amazon Mechanical Turk. We ask participants to find an instance of a particular action class in a $15$ minute video. We design a user interface that includes a time bar, which allows Turkers to navigate over the video quickly until the action is found. Typically, the action instances last three seconds, are sparsely localized, and at least one occurs in the video. 

We investigate two variants of the task: (i) a \emph{single class search} to find one instance of a given action class and (ii) a \emph{multiple class search}, which asks Turkers to find one instance from a larger set of action classes. By logging Turker interactions with the user interface, we measure the number of observed frames they require to find an action instance. As compared to \emph{single class search}, we find that Turkers observe $190\%$ and $210\%$ more frames when asked to find an action instance among $10$ and $20$ action classes, respectively. This observation motivates our intuition that action search can be performed more efficiently when the target task is ``simpler'' (\ie it incorporates a smaller number of action classes). See the \textbf{supplementary material} for more details about this experiment.

\noindent\textbf{Spotting vs. Localization.} \emph{Action spotting} is the process of finding any temporal occurrence of an action (\eg any frame inside a \emph{Long Jump} instance) while observing as little as possible from a video. In contrast, action localization focuses on pinpointing precise temporal extents of an action (\ie the exact start and end of a \emph{Long Jump} instance), which usually leads to fine-grained and exhaustive search mechanisms. To measure the effects of searching for fine-grained targets, we ask Turkers to find the starting time of a single target action in a video. We note that Turkers tend to first spot the action and then refine its temporal boundaries. Interestingly, while searching, Turkers perform \emph{three times} more search steps to refine the temporal boundaries of an action, as compared to the number of search steps needed for spotting the same action. This observation motivates our intuition that \emph{action spotting} can be performed more efficiently than action localization, especially when action instances are short in duration and sparse in frequency within a video. We partially attribute this behavior to the fact that determining precise temporal extents of some actions is ambiguous \cite{sigurdsson_2017}. See the \textbf{supplementary material} for more details about this experiment.

\noindent\textbf{Key Idea.} Searching for and localizing actions in video is a task humans can do efficiently. However, current automatic methods lack such ability. This shortcoming arises primarily from the fact that existing models are trained without an intelligent search mechanism. This is in part due to the limitations of existing datasets which only provide supervision about the action's temporal location in the video, while ignoring the entire process the annotator follows to find this action. To address these limitations, we collect two novel datasets of \emph{Human Searches}, where videos are annotated with the search steps a human follows to temporally spot/localize actions. We refer to such step sequences as \emph{search sequences}. Below, we describe these collected datasets (refer to the \textbf{supplementary material} for the annotation process details and additional statistics).

\textbf{(i) AVA searches (targets are actions):} We collect search sequences from the AVA v1.0 dataset \cite{ava}, which is composed of feature films and contains $192$ $15$-minute-long videos. We select $15$ action classes (out of the original $80$ actions) based on the two conditions: (1) the average action coverage is relatively small; and (2) the action class has at least $10$ videos in the training set. Based on these two conditions, actions such as \emph{talk to}, \emph{stand}, or \emph{watch (a person)} are discarded due to their extremely large coverage. Moreover, actions like \emph{shovel}, \emph{kick}, or \emph{exit} are discarded because they only have a few training samples. To gather the search sequences, we assign Turkers the task of spotting action instances in the AVA training videos. In other words, the Turker's task is to spot \emph{any} frame inside the temporal bounds of a given action. We only accept workers with more than $1000$ HITs submitted. In addition, we use the existing AVA dataset ground truth to filter out noisy annotations.
We use a total of $139$ AVA training videos and collect $3988$ search sequences. Notably, humans only observe a very small portion of the video (less than 1\%) before spotting the action.

\begin{figure*}[t!]
    \centering
    \begin{subfigure}{\linewidth}
        \includegraphics[width=1\linewidth]{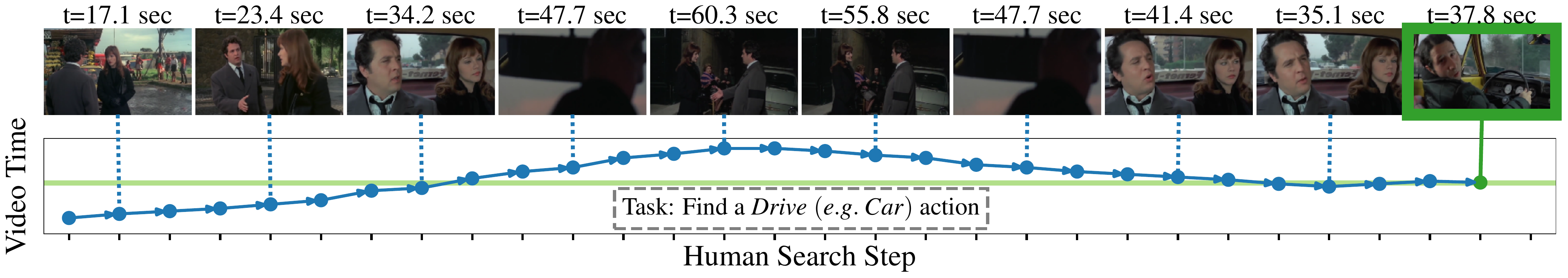}
    \end{subfigure}
    \begin{subfigure}{\linewidth}
        \includegraphics[width=1\linewidth]{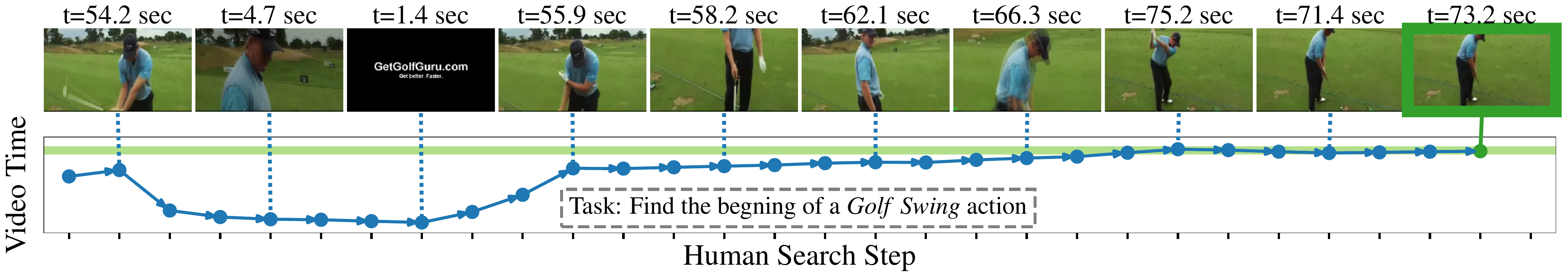}
    \end{subfigure}
    \caption{
        Illustration of two human search sequences from our \emph{Human Searches} dataset for an AVA \cite{ava} training video (first row) and a THUMOS14 \cite{thumos14} training video (second row). The shaded green areas are where the search targets occur. 
    }
    \label{fig:dataset_examples}
\end{figure*}

\textbf{(ii) THUMOS14 searches (targets are actions starting times):} 
We collect $1761$ search sequences from the training videos of THUMOS14 \cite{thumos14}, a large-scale dataset of untrimmed videos with $20$ different sports categories. We aim to use this searches dataset for \emph{action spotting} with the purpose of \emph{action localization}. Using the same collection process as in the \emph{AVA searches}, we ask Turkers to find an action's \emph{starting time}.
We choose to define the Turkers' task this way because defining the action's starting time is easier than defining its end time \cite{sigurdsson_2017}. 
Analyzing the collected data, we observe that humans find the actions in $6$ steps on average and define the starting points with an additional $16$ steps. This translates into observing only $5\%$ of the video. 
Figure \ref{fig:dataset_examples} shows two examples from the collected dataset. With its release, we hope \emph{Human Searches} will enable new research  directions in the video understanding field. 

\newcommand{\ahat}{\hat{\textbf{a}}}
\newcommand{\av}{\textbf{a}}
\newcommand{\bv}{\textbf{b}}
\newcommand{\cv}{\textbf{c}}
\newcommand{\dv}{\textbf{d}}
\newcommand{\uv}{\textbf{u}}
\newcommand{\vv}{\textbf{v}}
\newcommand{\x}{\textbf{x}}
\newcommand{\X}{\textbf{X}}
\newcommand{\y}{\textbf{y}}
\newcommand{\Y}{\textbf{Y}}
\newcommand{\z}{\textbf{z}}
\newcommand{\w}{\textbf{w}}
\newcommand{\W}{\textbf{W}}
\newcommand{\p}{\textbf{p}}
\newcommand{\q}{\textbf{q}}
\newcommand{\h}{\textbf{h}}
\newcommand{\A}{\textbf{A}}
\newcommand{\B}{\textbf{B}}
\newcommand{\D}{\textbf{D}}
\newcommand{\V}{\textbf{V}}
\newcommand{\U}{\textbf{U}}
\newcommand{\I}{\textbf{I}}
\newcommand{\PX}{\textbf{P}}
\newcommand{\mSigma}{\mathbf{\Sigma}}
\newcommand{\0}{\mathbf{0}}
\newcommand{\1}{\mathbf{1}}

\section{Proposed Action Search Model}\label{section:model}

\begin{figure*}[t]
    \centering
    \includegraphics[width=0.9\textwidth]{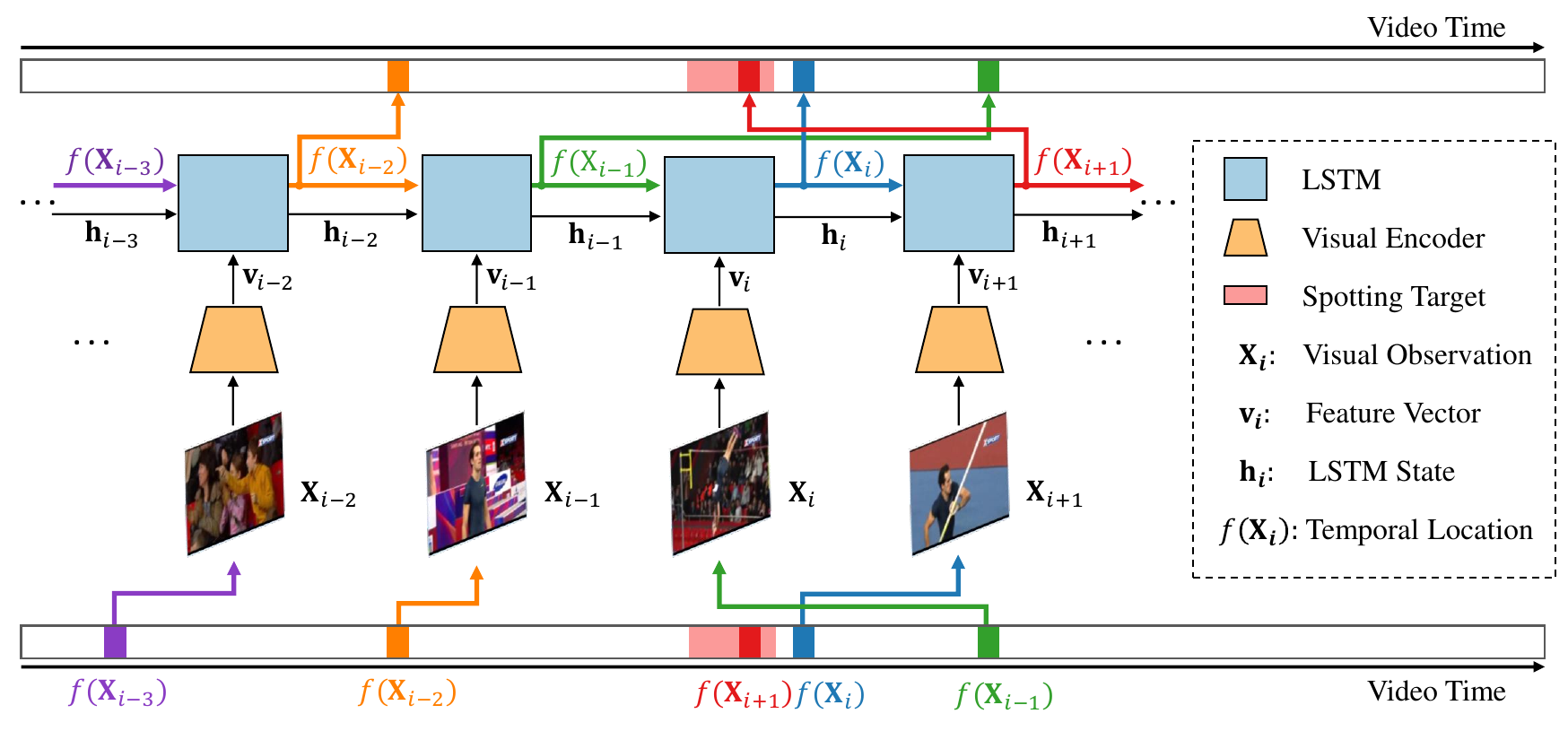}
    \caption{
        Our model harnesses the temporal context from its current location and the history of what it has observed to predict the next search location in the video. At each step, 
        (i) a visual encoder transforms the visual observation extracted from the model's current temporal location to a representative feature vector; 
        (ii) an LSTM consumes this feature vector plus the state and temporal location produced in the previous step;
        (iii) the LSTM outputs its updated state and the next search location; 
        (iv) the model moves to the new temporal location. 
    }
    \label{fig:pipeline}
\end{figure*}

In this section, we discuss the \emph{Action Search} model architecture, the approach used to train it, and some specific implementation details. 
Figure \ref{fig:pipeline} gives an overview of the main architecture of our approach. 

\subsection{Model Architecture} 
The input to \emph{Action Search} is a sequence of visual observations ($\X_1, \X_2, \dots, \X_n)$ and the output is a sequence of temporal locations $(f(\X_1), f(\X_2), \dots, f(\X_n))$, \ie a search sequence, produced by the search process. 
For a given video and at the $i^{\text{th}}$ search step, a visual encoder represents the observation $\X_i$ extracted from the model's current temporal location in the video by a feature vector $\vv_i$. 
We cast the search mechanism as a sequential decision making process modeled by an LSTM search network. This LSTM takes three inputs $(\h_{i-1}, f(\X_{i-1}), \vv_{i})$, where $\h_{i-1}$ is the LSTM state from the previous step (an aggregation of information from previous steps), $f(\X_{i-1})$ is the predicted temporal location from the previous step (the model's current location), and $\vv_i$ is the feature vector representing the current visual observation. 
Finally, a fully-connected layer transforms $\h_{i}$ (the updated state) to produce $f(\X_{i})$, the next location to search in the video.

\subsection{Learning to Search}
We employ a multi-layer LSTM network in training \emph{Action Search} to produce search sequences that align with human search sequences in the same video. Following the observations from the user study in Section \ref{section:action_searches}, \emph{action spotting} should be class-driven, and thus we train a separate LSTM for each action class. 

For a training video and at each search step, the network consumes visual observations at its current temporal location, along with the temporal location from the previous step.
After running for $n$ steps, the network produces a search sequence $(f(\X_1), f(\X_2), \dots, f(\X_n))$. Given the search sequence of a human annotator $(y_1, y_2, \dots, y_n)$ for the same video in our \emph{Human Searches} dataset, we compute the loss $L$ as the average Huber loss at each search step
\begin{align}
    L &=  \frac{1}{n}\sum_{i=1}^n H_{\delta}(y_i, f(\X_i)), \\
    H_{\delta}(y, f(\X)) &= 
      \begin{cases}
        \frac{1}{2}(y - f(\X))^2                   & \textrm{if } |y - f(\X)| \le \delta, \\
        \delta\, |y - f(\X)| - \frac{1}{2}\delta^2 & \textrm{otherwise.}
      \end{cases}
\end{align}
where $\delta > 0$. 
We choose the mean Huber loss over the mean squared loss, $\frac{1}{n}\sum_{i=1}^n (y_i - f(\X_i))^2$, in order to overcome the effects of outliers that might arise from the different Turkers searching the same video. Moreover, the mean Huber loss is convex and differentiable in a local neighborhood around its minimum, giving it an advantage over the mean absolute loss, $\frac{1}{n}\sum_{i=1}^n |y_i - f(\X_i)|$. 

\subsection{Implementation Details}
\noindent\textbf{Training Stage.} Although our pipeline is differentiable and can be trained end-to-end, we simplify and expedite training by fixing the visual encoder to precomputed ResNet-152 \cite{resnet} features (pretrained on ImageNet \cite{imagenet}) extracted from the average-pooling layer. We reduce the feature dimensionality to $512$  using PCA. We employ teacher forcing for trainig the LSTMs. For a stable training process, we represent each $f(\X_i)$ and $y_i$ as relative steps to the previous search location and normalize the ground truth output search sequence in a per-class fashion. Each LSTM network is trained using the Adam optimizer \cite{kingma_arxiv2014} with an exponential learning rate decay. We unroll the LSTM for a fixed number of steps, ranging from $2^2$ to $2^5$, for the backpropagation computation. To regularize the multi-layer LSTM network, we follow the RNN dropout techniques introduced by Pham \etal \cite{pham_arxiv2013} and Zaremba \etal \cite{zaremba_arxiv2014}. We set $\delta = 1$ in the Huber loss.

\noindent\textbf{Inference Stage.} Since videos typically contain multiple instances of the same action class, we initialize our model at multiple random points. Each search is run for a fixed number of steps. The number of initial points and the search sequence length are cross-validated per class using the validation subset. However, we prefer to launch many short search sequences as opposed to few long ones, since LSTM states tend to saturate and become unstable after a large number of iterations \cite{end_to_end_mem_networks}. Thus, one may view the \emph{Action Search} model as a random sampler with a smart local search: the first search steps are a random sampling of the video (exploration), while the later search steps are fine-grained steps (local search) that rely on the temporal context accumulated throughout the search. 

\subsection{Model Variants}

Here, we present different variants of \emph{Action Search} network structure and training. Experiments using these flavors show minimal effects on the final results.

\noindent\textbf{Training with Weighted Loss.} In order to put more emphasis on learning the steps towards the end of the ground truth search sequence, each term in the loss $L$ is weighted inversely proportional to how close the search step is to the target action instance. We consider this variant to give less weight to learning the first search steps a human annotator makes (seemingly random) before accumulating enough temporal context knowledge to guide subsequent search steps.

\noindent\textbf{Early Stopping Confidence Score.} Another flavor integrates a new module that consumes the LSTM state $\h$ along with the feature vector $\vv$ at each step and produces a probability $p$ to determine if the search sequence has reached the spotting target. To train this model variant, we assign a probability $p=1$ to all frames inside the spotting targets and $p=0$ elsewhere. We add to $L$ a new term that computes the softmax cross-entropy loss of this confidence score.

\section{Experiments}

\subsection{Action Search for Action Spotting}\label{section:exp_spotting}
In this subsection, we demonstrate that our model is able to 
mimic the human search process when spotting actions in the AVA dataset \cite{ava}.
We first introduce our experimental protocol including a brief description 
of the dataset and metric used. Then, we compare \emph{Action Search} against two spotting baseline models.

\noindent\textbf{Dataset.} We conduct this experiment using the AVA v1.0 dataset \cite{ava}, which is among the largest annotated action datasets and is composed of feature films. We pick $15$ action categories out of the original $80$ set of actions. 
We train our model using the collected \emph{AVA searches} from our \emph{Human Searches} dataset. 
However, we prune off the search sequences with less than $8$ steps.
Refer to Section \ref{section:action_searches} for a description of the \emph{AVA searches} and details about the action categories selection criteria. We evaluate our model on $35$ testing videos. We use AVA 3-second annotations to define temporal boundaries for each action instance.

\noindent\textbf{Metric.} We compare our model against other approaches according to the \emph{action spotting} metric, which we define to be the expected number of unique observations made per video until spotting an action. An action is spotted if the model lands \emph{anywhere} between its ground truth temporal bounds.

\noindent\textbf{Baseline Methods.} To demonstrate the effectiveness of our approach in learning to search efficiently, we consider two baseline models, \emph{Random Baseline} and \emph{Direction Baseline} (refer to the \textbf{supplementary material} for more baselines). 

\noindent\emph{Random Baseline}: This model picks both the search direction and step size randomly. In particular, if the current search step is at time $t$ of the video, it randomly picks between searching before $t$ in the interval $[0,t]$ or after $t$ in the interval $[t,d]$, where $d$ is the duration of the video. The model then picks the next search location randomly from a uniform distribution on the selected interval.
  
\noindent\emph{Direction Baseline}: This model picks the search direction using a trained \emph{direction network} and chooses the search step size randomly. In particular, if the current search step is at time $t$ of the video, its \emph{direction network} uses the visual observation of the current frame to decide between searching before or after $t$, and then picks the next search location randomly from a uniform distribution on the selected interval. 
The \emph{direction network} is based on a ResNet-152 \cite{resnet} architecture with a binary softmax classifier. To train this network, we annotated each video frame with the search direction that leads to the nearest ground truth instance boundary. The \emph{direction network} is class-specific and achieves an average of $95\%$ training and $91\%$ validation accuracy per class.

\begin{figure}[t!]
    \centering
    \includegraphics[width=.80\textwidth]{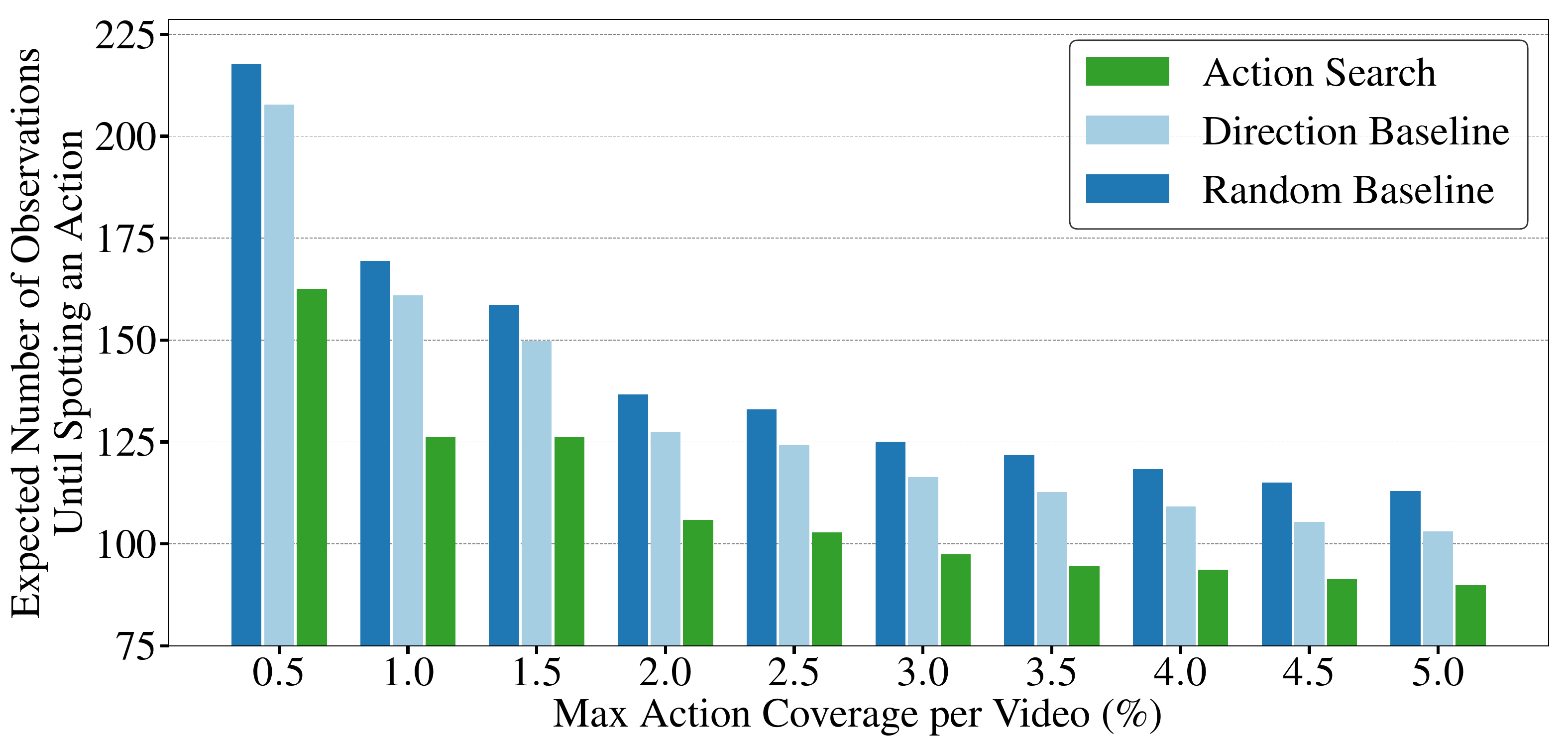}
    \caption{
        \emph{Action spotting} results for the AVA testing set for $1000$ independent search trials per video. We report the cumulative spotting metric results on videos with action coverage (\ie the percentage of video containing actions) $\le 5\%$. \emph{Action Search} takes $22\%$, $17\%$, and $13\%$ fewer observations than the \emph{Direction Baseline} on videos with at most $0.5\%$, $2.5\%$, and $5\%$ action coverage, respectively. 
    }
    \label{fig:action_spotting_vs_coverage}
\end{figure}

\noindent\textbf{Results and Analysis.} We run \emph{Action Search} and the two baseline methods for $1000$ independent search trials per test video to compute the \emph{action spotting} metric. At each trial, we initialize each model at a random temporal location in the video, and we reinitialize the search model if it fails to spot an action after $500$ steps. 
Figure \ref{fig:action_spotting_vs_coverage} shows the cumulative performance of all three models over videos with different action coverage (\ie the percentage of the video length that contains action instances).
Unlike both the \emph{Direction Baseline} and \emph{Random Baseline}, \emph{Action Search} is able to harness temporal context to expedite the search process and, as a result, takes on average $\mathbf{16.6}\%$ and $\mathbf{22.3}\%$ less observations, respectively, before successfully spotting actions.
Notably, there is a great performance difference between our model and the two baselines in videos with very sparse actions (\ie action coverage $\le 1\%$). We attribute this to the importance of temporal context when searching for sparse actions in video.

\subsection{Action Search for Action Localization}\label{section:exp_localization}
In this subsection, we combine \emph{Action Search} with an off-the-shelf action classifier to temporally localize human actions. Our approach achieves state-of-the-art detection performance on THUMOS14 \cite{thumos14}, one of the most challenging temporal action localization datasets, while observing only $\mathbf{17.3\%}$ of the video frames on average. We first explain how \emph{action spotting} is used as a precursor to temporal action localization, and then introduce our experimental protocol, including a brief description of the dataset and metric. Finally, we compare our method against state-of-the-art detectors and present an ablation study of our model.

\noindent\textbf{From Action Spotting to Localization.} We train \emph{Action Search} to spot the \emph{start} of actions using the \emph{THUMOS14 searches} dataset described in Section \ref{section:action_searches}. To define the spotted action length, we use a simple heuristic based on class-specific fixed prior lengths. Specifically, from an output search sequence $(f(\X_1), f(\X_2), \dots, f(\X_n))$, produced by our model, we generate the temporal segments
$\mathcal{Q} = \{\, [f(\X_i),f(\X_i) +  d] \mid i \in \{1, \dots, n\};~ d \in \mathcal{D} \,\}$,
where $\mathcal{D}$ is a set of class-specific fixed action duration priors precomputed from 
the validation videos. In pursuit of assigning a class label to each segment, we use a pretrained classifier to provide probabilistic action scores. Finally, we follow the standard practice of applying \emph{non-maximum-suppression} to remove duplicate detections.

\noindent\textbf{Prediction.}\label{section:prediction}
To reduce the number of search models we apply on a video to obtain its temporal predictions, we perform the following steps.
We uniformly sample a small random set of $N$ fixed-length segments from a given test video. These $N$ segments are then classified using an off-the-shelf action classifier in order to obtain the top-$k$ likely global class labels for the test video. To reduce  complexity, \emph{Action Search} then only runs the LSTM search models associated with these $k$ activity classes to produce a set of temporal search sequences. These sequences are then aggregated and transformed into a set of final temporal predictions using the class-specific duration priors described above. 

\noindent\textbf{Dataset.} We conduct our experiment using THUMOS14 \cite{thumos14}, one of the most popular datasets for temporal localization. It contains $200$ and $213$ videos for training and testing, respectively. To train our \emph{Action Search} model, we use the \emph{THUMOS14 searches} dataset described in Section \ref{section:action_searches}  (discarding the search sequences with less than $8$ search steps). Following the standard evaluation protocol, we evaluate our approach on the testing videos. We choose to do our experiments in THUMOS14 instead of other large-scale datasets such as Charades \cite{charades} or ActivityNet \cite{activitynet} due to the expensive costs of \emph{re-annotating} such datasets with search sequences. However, we provide an experiment in the \textbf{supplementary material} where we evaluate \emph{Action Search} (trained on THUMOS14) on the ActivityNet v1.2 validation videos with the same THUMOS14 classes.

\noindent\textbf{Metric.} We compare our model against other methods according to the mean Average Precision (mAP) and penalize duplicate detections. We report the mAP at multiple temporal Intersection-over-Union (tIoU) thresholds.

\noindent\textbf{Implementation Details.} We initialize \emph{Action Search} at random temporal locations in the video. We use the Res3D (+ S-CNN) classifier \cite{scnn,res3d} as an off-the-shelf pretrained action classifier to classify the $N$ random fixed-length segments, as well as, to classify the temporal action segments generated by our method. After cross-validating on the validation subset, we set $N=24$, as it achieves the highest average recall while maintaining a small number of segments. Empirically, we find setting $k=4$ provides the best trade-off between global video classification accuracy and number of search models to run.

\begin{table}[t!]
    \small

    {
    \renewcommand{\arraystretch}{1.2}

    \caption{
    	Temporal localization results (mAP at tIoU) on the THUMOS14 testing set. We assign `--' to unavailable mAP values. We report the average percentage of observed frames ($\mathbf{S}$) for each approach. 
    	\textbf{(a)} Comparison against state-of-the-art methods: Our method (\emph{Action Search + Priors} + Res3D + S-CNN) achieves state-of-the-art results while observing only $17.3\%$ of the video; 
    	\textbf{(b)} Video features effect: We compare C3D for \emph{Action Search} visual encoder + the C3D-based classifier from \cite{scnn} vs. ResNet for \emph{Action Search} visual encoder + the Res3D-based classifier from \cite{res3d};
    	\textbf{(c)} The trade-off between \emph{Action Search} training size and performance: mAP and \textbf{S} score improve as we increase the training size.
        }

    \begin{subtable}{.50\linewidth}
        \begin{flushleft}
            \caption{
            }
            \label{table:mAP-state-of-arts}
            \fontsize{7}{8.2}\selectfont
            \begin{tabular}{l || c c c c c | c}
            
                \hline
                
                \hline
                
                \multicolumn{1}{l ||}{}& \multicolumn{5}{c|}{\textbf{mAP at tIoU}} & \multicolumn{1}{l }{}\\
                \multicolumn{1}{l ||}{\textbf{Method}} & $0.3$ & $0.4$ & $0.5$ & $0.6$ & \multicolumn{1}{c|}{$0.7$} & \multicolumn{1}{c }{$\mathbf{S}$} \\
    
                \hline
                \hline
                
                \emph{Frame Glimpses} \cite{frameglimpses} & $36.0$ & $26.4$ & $17.1$ & -- & -- & $\underline{40}$
                \tablefootnote[1]{We assume each of the $20$ \emph{Frame Glimpses} \cite{frameglimpses} models observes $2\%$ of the video and report an upper-bound of $40\%$ frames observed.} \\ 
                Shou \etal \cite{scnn} & $36.3$ & $28.7$ & $19.0$ & -- & -- & $100$\\
                Shou \etal \cite{cdc} & $40.1$ & $29.4$ & $23.3$ & $13.1$ & $7.9$ & $100$\\
                Gao \etal \cite{turntap} & $44.1$ & $34.9$ & $25.6$ & -- & -- & $100$\\
                Dai \etal \cite{context_network_to_localize} & -- & $33.3$ & $25.6$ & $15.9$ & $9.0$ & $100$ \\
                Xu \etal \cite{rc3d} & $44.8$ & $35.6$ & $28.9$ & -- & -- & $100$\\
                Buch \etal \cite{sstad} & $45.7$ & -- & $29.2$ & -- & $9.6$ & $100$\\
                Zhao \etal \cite{ssn} & $\mathbf{51.9}$ & $41.0$ & $29.8$ & -- & --& $100$\\
                Gao \etal \cite{cbr} & $50.1$ & $\underline{41.3}$ & $\mathbf{31.0}$ & $\underline{19.1}$ & $\underline{9.9}$ & $100$\\

                \hline
                
                \emph{Res3D + S-CNN} \cite{res3d} & $40.6$ & $32.6$ & $22.5$ & $12.3$ & $6.4$ & $100$\\
                
                \textbf{Our method} & $\underline{51.8}$ & $\mathbf{42.4}$ & $\underline{30.8}$ & $\mathbf{20.2}$ & $\mathbf{11.1}$ & $\mathbf{17.3}$\\
                
                \hline
                
                \hline
                
            \end{tabular}    
        \end{flushleft}    
    \end{subtable}
\begin{minipage}{0.50\linewidth}
    
    \begin{subtable}{\linewidth}
        \begin{flushright}
            \caption{
            }
            \label{table:architecture_ablation}
            \fontsize{7.0}{7.2}\selectfont
            \begin{tabular}{c || c c c c c | c}
            
                \hline
                
                \hline
                
                \multicolumn{1}{c ||}{\textbf{Backbone}}& \multicolumn{5}{c|}{\textbf{mAP at tIoU}} & \multicolumn{1}{l }{}\\
                \multicolumn{1}{c ||}{\textbf{Arch.}} & $0.3$ & $0.4$ & $0.5$ & $0.6$ & \multicolumn{1}{c|}{$0.7$} & \multicolumn{1}{c }{$\mathbf{S}$} \\
    
                \hline
                \hline

                C3D & $43.7$ & $35.2$ & $23.9$ & $18.8$ & $4.3$ & $45.1$ \\
                ResNet & $51.8$ & $42.4$ & $30.8$ & $20.2$ & $11.1$ & $17.3$\\
                
                \hline
                
                \hline
                
            \end{tabular}    
        \end{flushright}    
    \end{subtable}

    \begin{subtable}{\linewidth}
        \begin{flushright}
            \caption{
            }
            \label{table:trainig_size_ablation}
            \fontsize{7.0}{7.2}\selectfont
            \begin{tabular}{c || c c c c c | c}
            
                \hline
                
                \hline
                
                \multicolumn{1}{c||}{\textbf{Training}}& \multicolumn{5}{c |}{\textbf{mAP at tIoU}} & \multicolumn{1}{l }{}\\
                \multicolumn{1}{c||}{\textbf{Size}} & $0.3$ & $0.4$ & $0.5$ & $0.6$ & \multicolumn{1}{c|}{$0.7$} & \multicolumn{1}{c }{$\mathbf{S}$} \\
    
                \hline
                \hline
                
                $0$\% \cite{res3d} & $40.6$ & $32.6$ & $22.5$ & $12.3$ & $6.4$ & $100$\\
                $25$\% & $41.1$ & $32.9$ & $22.4$ & $12.1$ & $6.4$ & $63.5$ \\
                $50$\% & $47.5$ & $38.3$ & $26.1$ & $16.6$ & $8.1$ & $34.4$ \\
                $75$\% & $50.2$ & $41.0$ & $29.5$ & $18.1$ & $9.4$ & $24.1$ \\
                $100$\% & $51.8$ & $42.4$ & $30.8$ & $20.2$ & $11.1$ & $17.3$\\

                \hline
                
                \hline
                
            \end{tabular}    
        \end{flushright}    
    \end{subtable}
    \end{minipage}

    }

\end{table}

\noindent\textbf{Results and Analysis.} In Table  \ref{table:mAP-state-of-arts}, we compare our localization approach with state-of-the-art techniques. We assess methods in terms of mAP and the average percentage of observed frames ($\mathbf{S}$). Our approach achieves state-of-the-art detection performance while observing much less of the video. For instance, at $0.5$ tIoU, we achieve a competitive $30.8\%$ mAP (compared to \cite{cbr}'s $31.0\%$ mAP) and observe on average $17.3\%$ of each video. To understand the strengths of our method, we break down the results and discuss the following findings:

\textbf{(1)} Our method outperforms its baseline (Res3D + S-CNN \cite{res3d}) by $8.3\%$ mAP (0.5 tIoU). This baseline uses the same classifier \emph{Action Search} uses, but its action proposals are generated by a CNN. We attribute this improvement to our approach's ability to discard irrelevant portions of the video, which allows the detector to prune false positive detections. This finding indicates the importance of \emph{Action Search}  as a precursor to localization.

\textbf{(2)} Our method outperforms \emph{Frame Glimpses} \cite{frameglimpses} in both mAP and $\mathbf{S}$. We observe a $13.7\%$ mAP ($0.5$ tIoU) improvement and a $22.7\%$ reduction in $\mathbf{S}$. We attribute the improvement on $\mathbf{S}$ to the strategy we follow to train \emph{Action Search}. Instead of relying on a reinforcement policy to explore the video, as \emph{Frame Glimpses} does, we learn to search by imitating humans. Thus, we argue that their search policy is unable to learn temporal context reasoning as well as \emph{Action Search}. This finding justifies the need for the \emph{Human Searches} dataset (Section \ref{section:action_searches}) and the supervised strategy we follow to train our model.

\textbf{(3)} Although it uses a naive approach to detect actions, our method surpasses state-of-the-art approaches \cite{sstad,cbr,ssn}. Opting for simplicity, we build our detector by combining \emph{Action Search} with a pretrained off-the-shelf classifier. This straightforward approach allows us to achieve state-of-the-art results while only observing $17.3\%$ of video frames. We argue further improvements can be obtained by combining \emph{Action Search} with sophisticated models such as \cite{cbr,ssn}.

\noindent\textbf{Ablation Study.} We investigate two aspects of our model: \textbf{(i)} the backbone architecture for the \emph{Action Search} visual encoder and the off-the-shelf classifier (refer to Table \ref{table:architecture_ablation}), and \textbf{(ii)} the trade-off between \emph{Action Search} training set size and performance in terms of mAP and \textbf{S} (refer to Table \ref{table:trainig_size_ablation}). 

\textbf{(i)} Table \ref{table:architecture_ablation} shows using a C3D architecture (\ie C3D for \emph{Action Search} visual encoder + the C3D-based classifier from \cite{scnn}) improves the performance of the baseline \cite{scnn} by $4.9\%$ mAP ($0.5$ tIoU) while observing $54.9\%$ less frames. However, a ResNet backbone architecture gives better results in both mAP and \textbf{S}. We attribute this to the facts that ResNet offers a richer feature space, the off-the-shelf classifier from \cite{res3d} is more sophisticated compared to the classifier from \cite{scnn}, and ResNet uses 1 frame per observation while C3D needs 16 frames.

\textbf{(ii)} \emph{Action Search} is trained on \emph{THUMOS14 searches} dataset, which contains on average $8.8$ human searches per video. Training on more human searches would lead to a better performing model, but collecting human searches for \emph{existing} datasets is expensive (although \emph{free} for new datasets). Nonetheless, we observe in Table \ref{table:trainig_size_ablation} that training \emph{Action Search} on half of the \emph{THUMOS14 searches} dataset (\ie $4.4$ human searches per video), the model improves the detection performance of \cite{res3d} by $3.6\%$ mAP ($0.5$ tIoU) while observing only \textbf{S}=$34.4\%$ of the frames. Notably, training on as little as $2.2$ human searches per video cuts the percentage of observed frames by $36.5\%$ while keeping a similar mAP performance. In general, as we decrease the training set size, the efficiency of the search degrades (\textbf{S} increases) since the model is not exposed to as many human searches variations. These findings further justify the need for the \emph{Human Searches} dataset and the supervised strategy we follow to train our model.

\begin{figure}[t!]
    \centering
    \begin{subfigure}{0.49\linewidth}
        \includegraphics[width=1\linewidth]{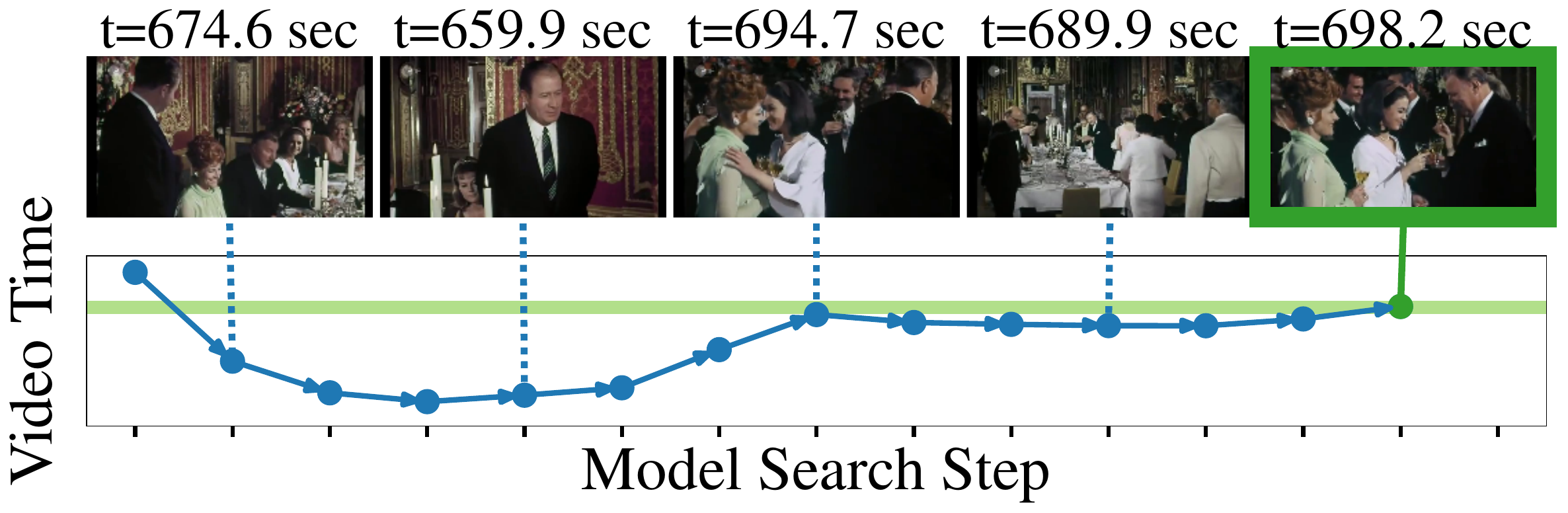}
    \end{subfigure}
    \begin{subfigure}{0.49\linewidth}
        \includegraphics[width=1\linewidth]{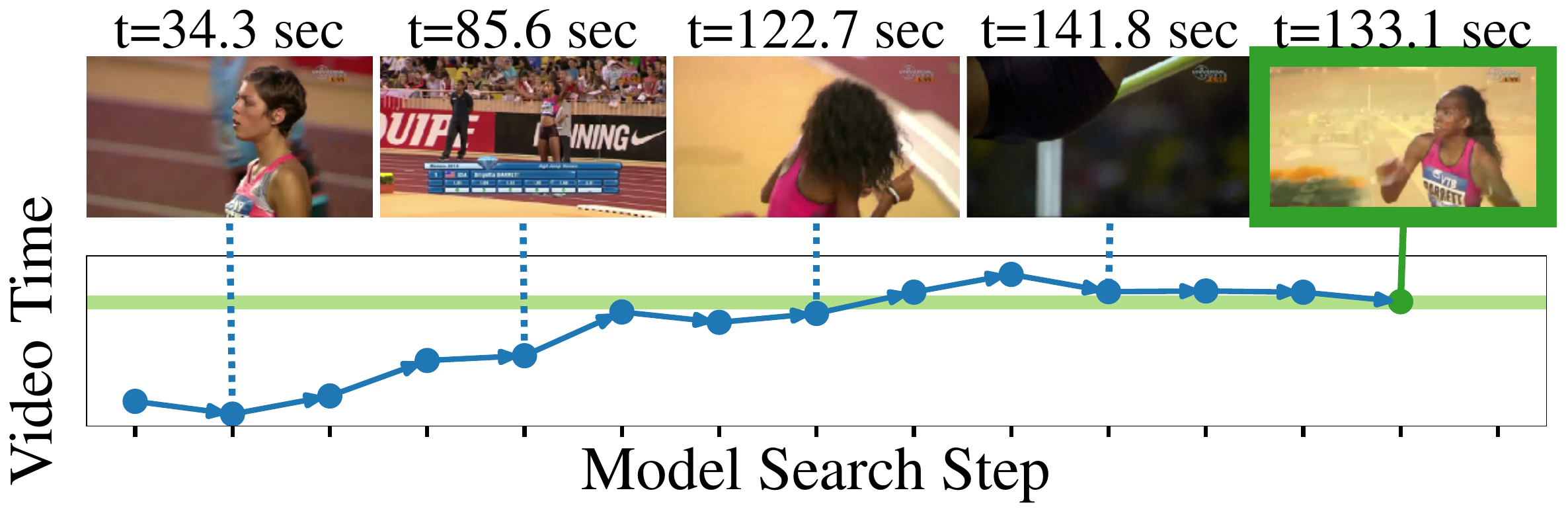}
    \end{subfigure}
    
    \begin{subfigure}{0.49\linewidth}
        \includegraphics[width=1\linewidth]{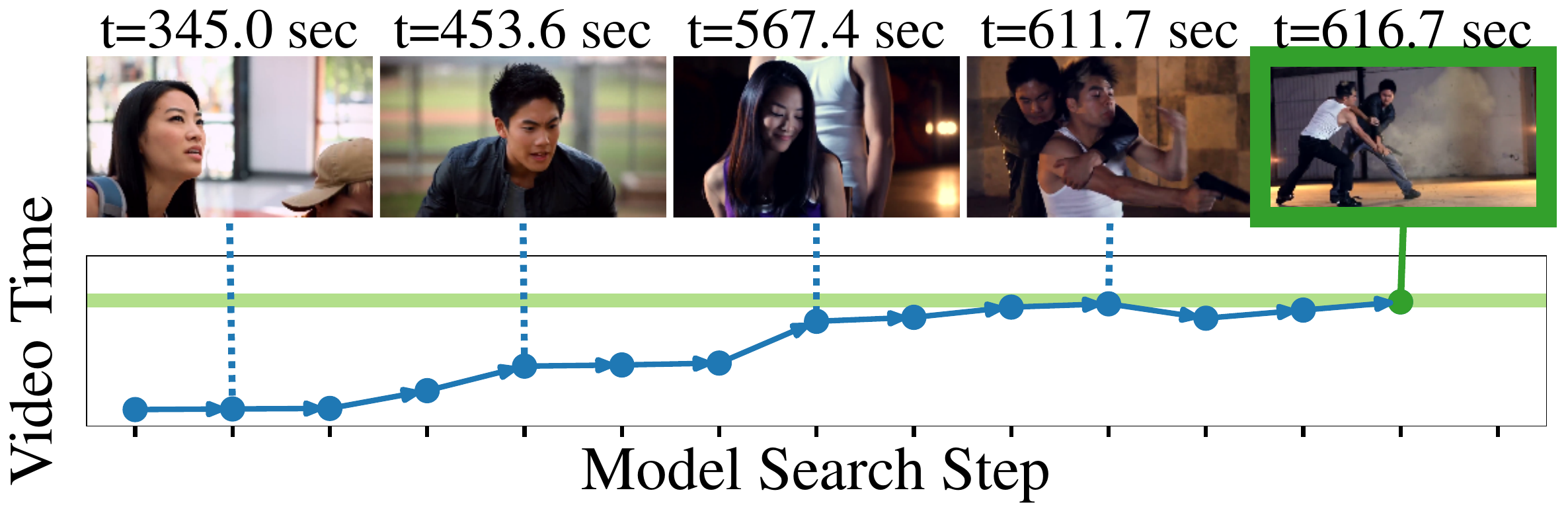}
    \end{subfigure}
    \begin{subfigure}{0.49\linewidth}
        \includegraphics[width=1\linewidth]{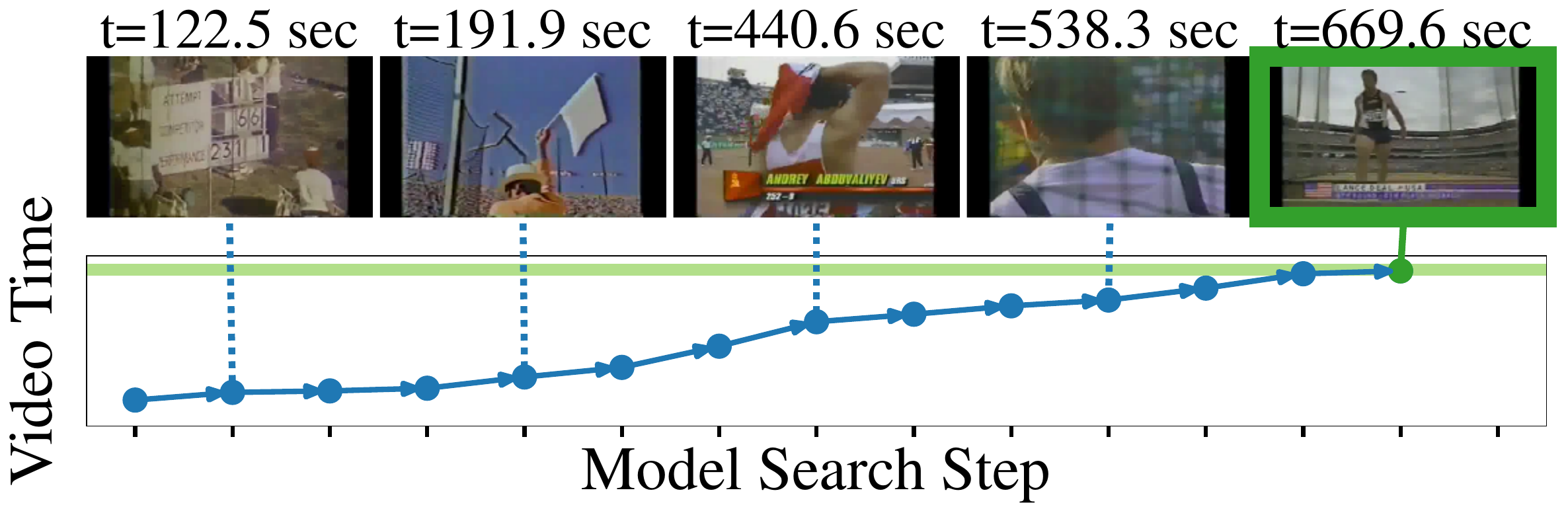}
    \end{subfigure}    

    \begin{subfigure}{0.49\linewidth}
        \includegraphics[width=1\linewidth]{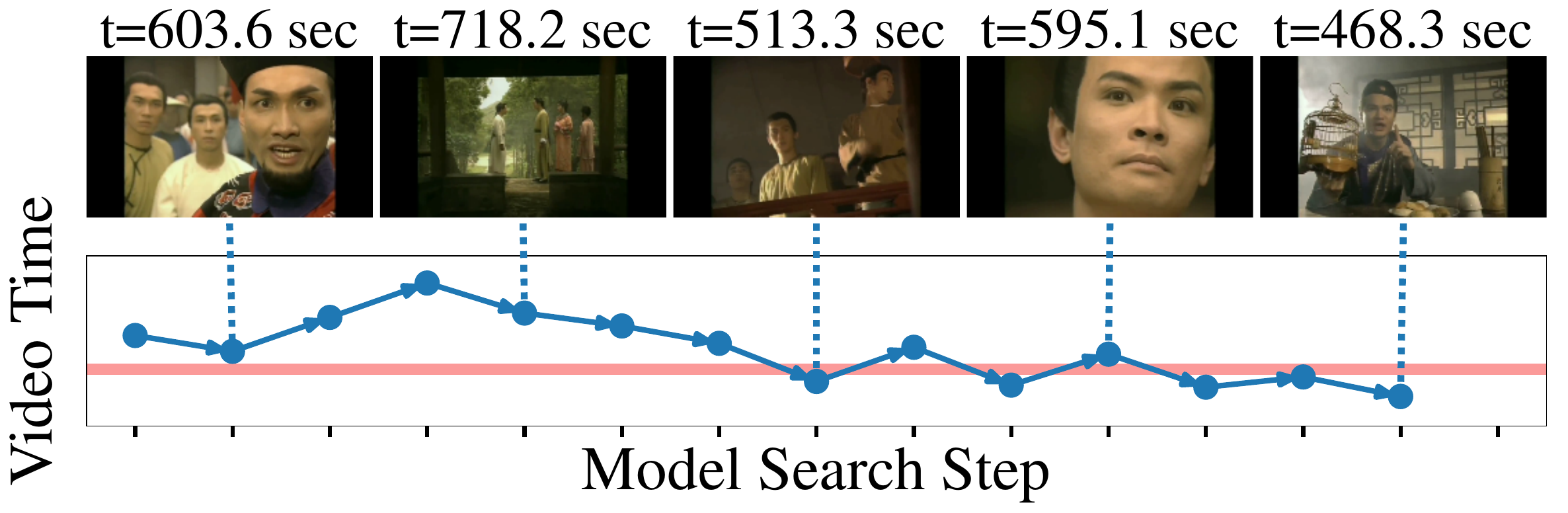}
    \end{subfigure} 
    \begin{subfigure}{0.49\linewidth}
        \includegraphics[width=1\linewidth]{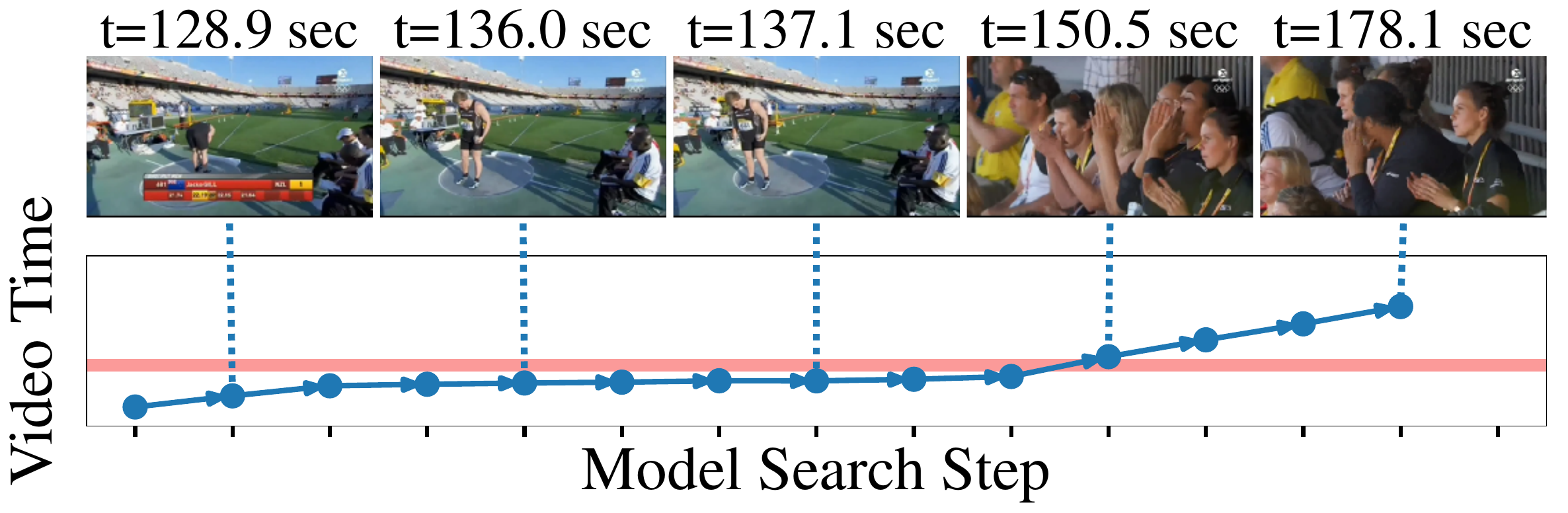}
    \end{subfigure}

    \caption{
        Qualitative search sequences produced by \emph{Action Search}. The left column corresponds to AVA \cite{ava} testing videos, and the right column corresponds to THUMOS14 \cite{thumos14} testing videos. The top two rows depict examples when our model successfully spots the target action location (in green). The last row illustrate failure cases, \ie when the action location (in red) is not spotted exactly. We observe that \emph{Action Search} uses temporal context to reason about where to search next. In failure cases, we notice that our model often oscillates around actions without spotting frames within the exact temporal location.
    }
    \label{fig:qualitative}
\end{figure}

\subsection{Spotting at a Glance}
Figure \ref{fig:qualitative} depicts qualitative search sequences of our \emph{Action Search} model, which exploits  temporal context to spot the actions quickly. For example, it uses information about scenes/concepts such as gala dinner and fights to spot actions such as \emph{clink glass} and \emph{shoot}, respectively (top two rows, left column). Furthermore, when spotting the start times of actions, our model seems to understand the inherent temporal structures of actions. \emph{Action Search} surprisingly understands when an action finishes and is able to rewind (jump back) and spot the beginning of the action (top row, right column). Typical failure cases (actions are not spotted) occur when the search sequences oscillate around the target action but they miss its exact location (last row, left column). Additionally, \emph{Action Search} may fail when action boundaries are ambiguous. When our model finds content visually similar to the target action, it remains static for a few search steps and then decides to explore the video further (last row, right column).

\section{Conclusion}
In this paper, we introduced \emph{Action Search}, a new learning model to imitate how humans search for actions in videos, and a new dataset called \emph{Human Searches} to train such model. Extensive experiments demonstrated that \emph{Action Search} produces reliable action detections. We plan to release our \emph{Human Searches} dataset to the vision community and expect that further works can extend the use of search processes for action detection and other applications.\\

\noindent\textbf{Acknowledgments.}
This publication is based upon work supported by the King Abdullah University of Science and Technology (KAUST) Office of Sponsored Research (OSR) under Award No. OSR-CRG2017-3405.

\clearpage

\bibliographystyle{splncs04}
\bibliography{mybib}

\end{document}